# A hybrid spatial–temporal deep learning architecture for lane detection

Yongqi Dong[1] | Sandeep Patil[2] | Bart van Arem[1] | Haneen Farah[1]

[1] Department of Transport and Planning, Faculty of Civil Engineering and Geosciences, Delft University of Technology, Delft, The Netherlands

[2] Faculty of Mechanical, Maritime and Materials Engineering, Delft University of Technology, Delft, The Netherlands

**Correspondence**
Haneen Farah, Department of Transport and Planning, Faculty of Civil Engineering and Geosciences, Delft University of Technology, Delft, The Netherlands.
Email: h.farah@tudelft.nl

**Funding information**
Applied and Technical Sciences (TTW), a subdomain of the Dutch Institute for Scientific Research (NWO), Grant/Award Number: 17187

**Abstract**

Accurate and reliable lane detection is vital for the safe performance of lane-keeping assistance and lane departure warning systems. However, under certain challenging circumstances, it is difficult to get satisfactory performance in accurately detecting the lanes from one single image as mostly done in current literature. Since lane markings are continuous lines, the lanes that are difficult to be accurately detected in the current single image can potentially be better deduced if information from previous frames is incorporated. This study proposes a novel hybrid spatial–temporal (ST) sequence-to-one deep learning architecture. This architecture makes full use of the ST information in multiple continuous image frames to detect the lane markings in the very last frame. Specifically, the hybrid model integrates the following aspects: (a) the single image feature extraction module equipped with the spatial convolutional neural network; (b) the ST feature integration module constructed by ST recurrent neural network; (c) the encoder–decoder structure, which makes this image segmentation problem work in an end-to-end supervised learning format. Extensive experiments reveal that the proposed model architecture can effectively handle challenging driving scenes and outperforms available state-of-the-art methods.

## 1 | INTRODUCTION

The interest in developing automated driving functionalities, and in the end, fully automated vehicles, has been increasing vastly over the last decade. The safety of these automated functionalities is a crucial element and a priority for academic researchers, manufacturers, policymakers, and their potential future users. Automated driving requires a full understanding of the environment around the automated vehicle through its sensors. Vision-based methods have lately been boosted by advancements in computer vision and machine learning. Regarding environmental perception, camera-based lane detection is important, as it allows the vehicle to position itself within the lane. This is also the foundation of most lane-keeping assistance and lane departure warning systems (Andrade et al., 2019; Bar Hillel et al., 2014; W. Chen et al., 2020; Liang et al., 2020; Xing et al., 2018).

Traditional vision-based lane-detection methods rely on hand-crafted low-level features (e.g., color, gradient, and ridge features) and usually work in a four-step procedure, that is, image pre-processing, feature extraction, line detection and fitting, and post-processing (Bar Hillel et al., 2014; Haris & Glowacz, 2021). Traditional computer vision techniques, for example, Inverse Perspective Mapping (Aly, 2008; Wang et al., 2014), Hough transform (Berriel et al., 2017; Jiao et al., 2019; Zheng et al., 2018), Gaussian filters (Aly, 2008; Sivaraman and Trivedi, 2013; Wang et al., 2012), and Random Sample Consensus (RANSAC) (Aly, 2008; Choi et al., 2018; Du et al., 2018; Guo et al., 2015; Lu et al., 2019), are usually adopted in the 4-step procedure. The problems of traditional methods are: (a) hand-crafted features are cumbersome to manage and not always useful, suitable, or powerful; and (b) the detection results are always based on one single image. Thus, the detection accuracies are relatively not high.

During the last decade, with the advancements in deep learning algorithms and computational power, many deep neural network based methods have been developed for lane detection with good performance. There are generally two dominant approaches (Tabelini et al., 2020b), i.e., (1) segmentation-based pipeline (Kim and Park, 2017; Ko et al., 2020; T. Liu et al., 2020; Pan et al., 2018; Zhang et al., 2021; Zou et al., 2020), in which predictions are made on the per-pixel basis, classifying each pixel as either lane or not; (2) the pipeline using row-based prediction (Hou et al., 2020; Qin et al., 2020; Yoo et al., 2020), in which the image is split into a (horizontal) grid and the model predicts the most probable location to contain a part of a lane marking in each row. Recently, Liu et al. (2021) summarized two additional categories of deep learning based lane detection methods, i.e., the anchor-based approach (Chen et al., 2019; Li et al., 2020; Tabelini et al., 2020b; Xu et al., 2020), which focuses on optimizing the line shape by regressing the relative coordinates with the help of predefined anchors, and the parametric prediction based method which directly outputs parametric lines expressed by curve equation (R. Liu et al., 2020; Tabelini et al., 2020a). Apart from these dominant approaches, some other less common methods were proposed recently. For instance, Lin et al. (2020) fused the adaptive anchor scheme (designed by formulating a bilinear interpolation algorithm) aided informative feature extraction and object detection into a single deep convolutional neural network for lane detection from a top-view perspective. Philion (2019) developed a novel learning-based approach with a fully convolutional model to decode the lane structures directly rather than delegating structure inference to post-processing, plus an effective approach to adapt the model to new contexts by unsupervised transfer learning.

Similar to traditional vision-based lane-detection methods, most available deep learning models utilize only the current image frame to perform the detection. Until very recently, a few studies have explored the combination of convolutional neural network (CNN) and recurrent neural network (RNN) to detect lane markings or simulate autonomous driving using continuous driving scenes (Chen et al., 2020; Zhang et al., 2021; Zou et al., 2020). However, the available methods do not take full advantage of the essential properties of the lane being long continuous solid or dashed line structures. Also, they do not yet make the utmost of the spatial-temporal information together with correlation and dependencies in the continuous driving frames. Thus, for certain extremely challenging driving scenes, their detection results are still unsatisfactory.

In this paper, lane detection is treated as a segmentation task, in which a novel hybrid spatial-temporal sequence-to-one deep learning architecture is developed for lane detection through a continuous sequence of images in an end-to-end approach. To cope with challenging driving situations, the hybrid model takes multiple continuous frames of an image sequence as inputs, and integrates the single image feature extraction module, the spatial-temporal feature integration module, together with the encoder-decoder structure to make full use of the spatial-temporal information in the image sequence. The single image feature extraction module utilizes modified common backbone networks with embedded spatial convolutional neural network (SCNN) (Pan et al., 2018) layers to extract the features in every single image throughout the continuous driving scene. SCNN is powerful in extracting spatial features and relationships in one single image, especially for long continuous shape structures. Next, the extracted features are fed into spatial-temporal recurrent neural network (ST-RNN) layers to capture the spatial-temporal dependencies and correlations among the continuous frames. An encoder-decoder structure is adopted with the encoder consisting of SCNN and several fully-convolution layers to downsample the input image and abstract the features, while the decoder, constructed by CNNs, upsample the abstracted outputs of previous layers to the same size as the input image. With the labelled ground truth of the very last image in the continuous frames, the model training works in an end-to-end way as a supervised learning approach. To train and validate the proposed model on two large-scale open-sourced datasets, i.e., tvtLANE (Zou et al., 2020) and TuSimple, a corresponding training strategy has been also developed. To summarize, the main contributions of this paper lie in:

• A hybrid spatial-temporal sequence-to-one deep neural network architecture integrating the advantages of the encoder-decoder structure, SCNN embedded single image feature extraction module, and ST-RNN module, is proposed;

• The proposed model architecture is the first attempt that tries to strengthen both spatial relation feature extraction in every single image frame and spatial-temporal correlation together with dependencies among continuous image frames for lane detection;

• The implementation utilized two widely used neural network backbones, i.e., UNet (Ronneberger et al., 2015) and SegNet (Badrinarayanan et al., 2017) and included extensive evaluation experiments on commonly used datasets, demonstrating the effectiveness and strength of the proposed model architecture;

• The proposed model can tackle lane detection in challenging scenes such as curves, dirty roads, serious vehicle occlusions, etc., and outperforms all the available state-of-the-art baseline models in most cases with a large margin.

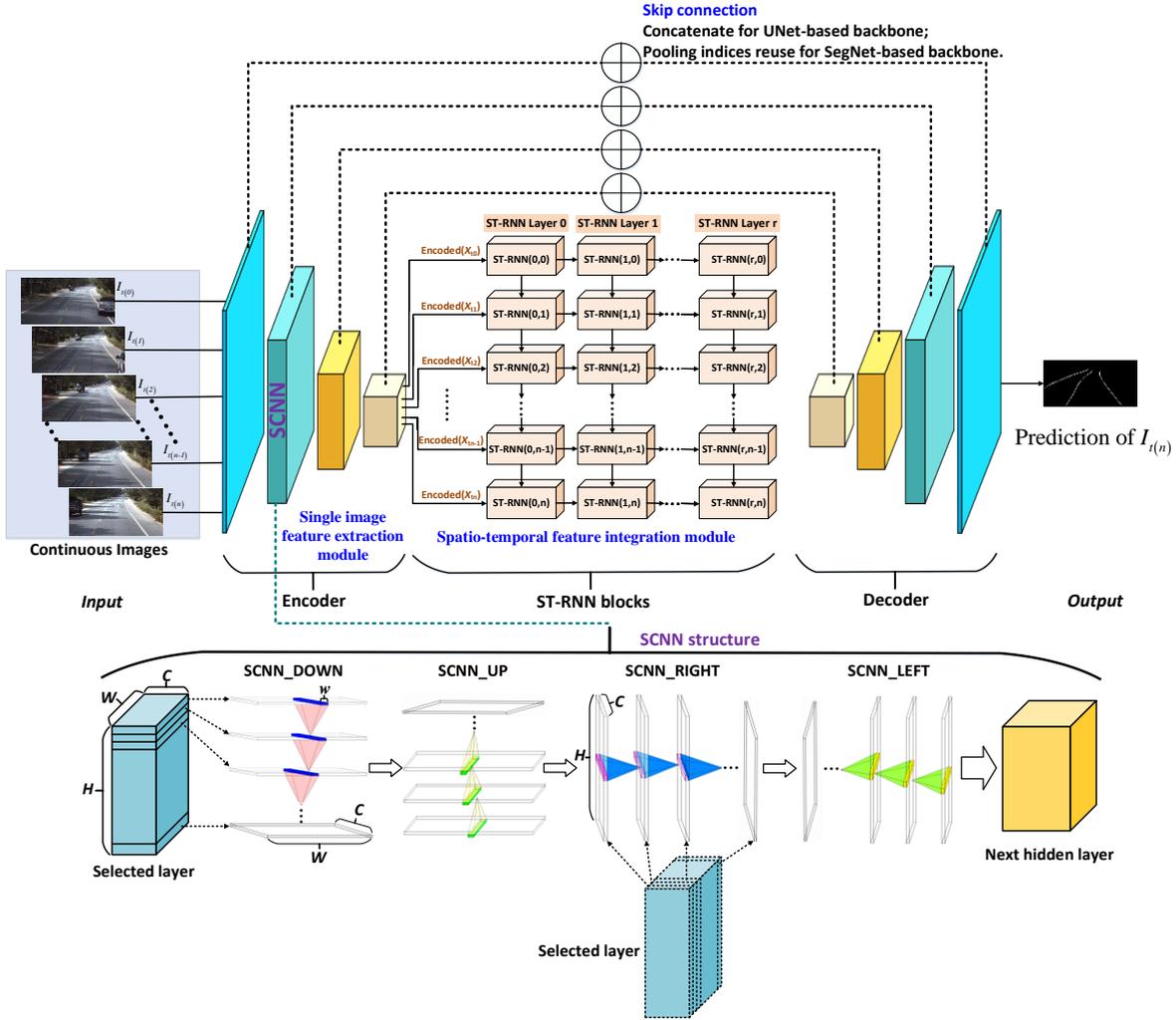

**FIGURE 1.** The architecture of the proposed model.

- Under the proposed architecture, the light version model variant can achieve beyond state-of-the-art performance while using fewer parameters.

## 2 PROPOSED METHOD

Although many sophisticated methods have been proposed for lane detection, most of the available methods use only one single image resulting in unsatisfactory performance under some extremely challenging scenarios, e.g., dazzle lighting, and serious occlusion. This study proposes a novel hybrid spatial-temporal sequence-to-one deep neural network architecture for lane detection. The architecture was inspired by: (a) the successful precedents of hybrid deep neural network architectures which fuse CNN and RNN to make use of by: (a) the successful precedents of hybrid deep neural network architectures which fuse CNN and RNN to make use of information in continuous multiple frames (Zhang et al., 2021; Zou et al., 2020); (b) the domain prior knowledge that traffic lanes are long continuous shape line structure with strong spatial relationship. The architecture integrates two modules utilizing two distinctive neural networks with complementary merits, i.e., SCNN and convolutional Long Short Term Memory (ConvLSTM) neural network, under an end-to-end encoder-decoder structure, to tackle lane detection in challenging driving scenes.

### 2.1 Overview of the proposed model architecture

The proposed deep neural network architecture adopts a sequence-to-one end-to-end encoder-decoder structure as shown in Figure 1.

Here "sequence-to-one" means that the model gets a sequence of multi images as input and outputs the detection result of the last image (please note that essentially the model is still utilizing sequence-to-sequence neural networks); "end-to-end" means that the learning algorithm goes directly from the input to the desired output, which refers to the lane detection result in this paper, bypassing the intermediate states (Levinson et al., 2011; Neven et al., 2017); the encoder-decoder structure is a modular structure that consists of an encoder network and a decoder network, and is often employed in sequence-to-sequence tasks, such as language translation e.g., (Sutskever et al., 2014), and speech recognition e.g., (Wu et al., 2017). Here, the proposed model adopts encoder CNN with SCNN layers and decoder CNN using fully convolutional layers. The encoder takes a sequence of continuous image frames, i.e., time-series-images, as input and abstracts the feature map(s) in smaller sizes. To make use of the prior knowledge that traffic lanes are solid- or dashed- line structures with a continuous shape, one special kind of CNN, i.e., SCNN, is adopted after the first CNN hidden layer. With the help of SCNN, spatial features and relationships in every single image will be better extracted. Following this, the extracted feature maps of the continuous frames, constructed in a time-series manner, will be fed to ST-RNN blocks for sequential feature extraction and spatial-temporal information integration. Finally, the decoder network upsamples the abstracted feature maps obtained from the ST-RNN and decodes the content to the original input image size with the detection results. The proposed model architecture is implemented with two backbones, UNet (Ronneberger et al., 2015) and SegNet (Badrinarayanan et al., 2017). Note, in the UNet based architecture, similar to (Ronneberger et al., 2015), the proposed model employs the skip connection between the encoder and decoder phase by concatenating operation to reuse features and retain information from previous encoder layers for more accurate predictions; while in the SegNet based networks, at the decoder stage, similar to (Badrinarayanan et al., 2017), the proposed model reuses the pooling indices to capture, store, and make use of the vital boundary information in the encoder feature maps. The detailed network implementation is elaborated in the remaining parts of Section 2.

## 2.2 Network design

*1) End-to-end encoder-decoder*: Regarding lane detection as an image segmentation problem, the encoder-decoder structure based neural network can be implemented and trained in an end-to-end way. Inspired by the excellent performance of CNN-based encoder-decoder for image semantic-segmentation tasks in various domains (Badrinarayanan et al., 2017; Wang et al., 2020; Yasrab et al., 2017), this study also adopts the "symmetrical" encoder-decoder as the main backbone structure. Convolution and pooling operations are employed to extract and abstract the features in every image in the encoder stage; while in the decoder subset, the inverted convolution and upsampling operation are adopted to grasp the extracted high-order features and construct the outputs layer by layer with regards to the targets. By setting the output target size the same as the input image size, the whole network can work in an end-to-end approach. In the implementation, two widely used backbones, U-Net and Seg-Net, are adopted. To better extract and make use of the spatial relations in every image frame, the SCNN layer is introduced in the encoder part of the single image feature extraction module. Furthermore, to excavate and make use of the spatial-temporal correlations and dependencies among the input continuous image frames, ST-RNN blocks are embedded in the middle of the encoder-decoder networks.

*2) SCNN*: The Spatial Convolutional Neural Network (SCNN) was first proposed by Pan et al. (2018). The "spatial" here means that the specially designed CNN can propagate spatial information via slice-by-slice message passing. The detailed structure of SCNN is demonstrated in the bottom part of Figure 1.

SCNN can propagate the spatial information in one image through four directions, as shown with the suffix "DOWN", "UP", "RIGHT", "LEFT" in Figure 1, which denotes downward, upward, rightward, and leftward, respectively. Take the "SCNN_DOWN" module for an example, considering that SCNN is adopted on a three dimensional tensor of size $C \times W \times H$, where in the lane detection task, $C$, $W$, and $H$ denote the number of channels, image (or its feature map) width, and heights respectively. For SCNN_D, the input tensor would be split into $H$ slices, and the first slice will then be sent into a convolution operation layer with $C$ kernels of size $C \times w$, in which $w$ is the kernel width. Different from the traditional CNN in which the output of one convolution layer is introduced into the next layer directly, in SCNN_D the output is added to the next adjacent slice to produce a new slice, and iteratively to the next convolution layer continuing until the last slice in the selected direction is updated. The convolution kernel weights are shared throughout all slices, and the same mechanism works for other directions of SCNNs.

With the above properties, SCNN has demonstrated its strengths in extracting spatial relationships in the image, which makes it suitable for detecting long continuous shape structures, e.g., traffic lanes, poles, and walls (Pan et al., 2018). However, using only one image to do the detection, SCNN still could not produce satisfying performance under extremely challenging conditions. And that is why a sequence-to-one architecture with continuous image frames as inputs and ST-RNN blocks to capture the spatial-temporal correlations in the continuous frames is proposed in this paper.

*3) ST-RNN module*: In this proposed framework, the multiple continuous frames of images are modelled as "image-time-series" inputs. To capture the spatial-temporal dependencies and correlations among the image-time-series, the ST-RNN module is embedded in the middle of the encoder-decoder structure, which takes over the output extracted features of the encoder as its input and outputs the integrated spatial-temporal information to the decoder.

Various versions of RNNs have been proposed, e.g., Long Short Term Memory (LSTM) together with its multivariate version, i.e., fully connected LSTM (FC-LSTM), and Gated Recurrent Unit (GRU), to tackle time-series data in different application domains. In this paper, two state-of-the-art RNN networks, i.e., ConvLSTM (Shi et al., 2015) and Convolutional Gated Recurrent Unit (ConvGRU) (Ballas et al., 2016), are employed. These models, considering their abilities in spatial-temporal feature extraction, generally outperform other traditional RNN models.

A general critical problem for the vanilla RNN model is the gradients vanishing (Hochreiter and Schmidhuber, 1997; Pascanu et al., 2013; Ribeiro, 2020). For this, LSTM introduces memory cells and gates to control the information flow to trap the gradient preventing it from vanishing during the back-propagation. In LSTM, the information of the new time-series inputs will be accumulated to the memory cell $C_t$ if the input gate $i_t$ is on. In contrast, if the information is not "important", the past cell status $C_{t-1}$ could be "forgotten" by activating the forget gate $f_t$. Also, there is the output gate $o_t$ which decides whether the latest cell output $C_t$ will be propagated to the final state $\mathcal{H}_t$. The traditional FC-LSTM contains too much redundancy for spatial information, which makes it time-consuming and computational-expensive. To address this, the ConvLSTM (Shi et al., 2015) is selected to build the ST-RNN block of the proposed framework. In ConvLSTM, the convolutional structures and operations are introduced in both the *input-to-state* and *state-to-state* transitions to do spatial information encoding, which also alleviates the problem of time- and computation-consuming.

The key formulation of the ConvLSTM is shown by equations (1)-(5), where $\odot$ denotes the Hadamard product, $*$ denotes the convolution operation, $\sigma(\cdot)$ represents the sigmoid function, and $\tanh(\cdot)$ represents the hyperbolic tangent function; $X_t$, $C_t$, and $\mathcal{H}_t$ are the input (i.e., the extracted features from the encoder in the proposed framework), memory cell status, and output at time $t$; $i_t$, $f_t$, and $o_t$ are the function values of the input gate, forget gate, and output gate, respectively; $W$ denotes the weight matrices, whose subscripts indicate the two corresponding variables are connected by this matrix. For instance, $W_{xc}$ is the weight matrix between the input extracted features $X_t$ and the memory cell $C_t$; $'b'$s are biases of the gates, e.g., $b_i$ is the input gate's bias.

$$i_t = \sigma(W_{xi} * X_t + W_{hi} * \mathcal{H}_{t-1} + W_{ci} \odot C_{t-1} + b_i) \quad (1)$$
$$f_t = \sigma(W_{xf} * X_t + W_{hf} * \mathcal{H}_{t-1} + W_{cf} \odot C_{t-1} + b_f) \quad (2)$$
$$C_t = f_t \odot C_{t-1} + i_t \odot \tanh(W_{xc} * X_t + W_{hc} * \mathcal{H}_{t-1} + b_c) \quad (3)$$
$$o_t = \sigma(W_{xo} * X_t + W_{ho} * \mathcal{H}_{t-1} + W_{co} \odot C_t + b_o) \quad (4)$$
$$\mathcal{H}_t = o_t \odot \tanh(C_t) \quad (5)$$

The ConvGRU (Ballas et al., 2016) further lightens the computational complexity by reducing a gate structure but could perform similarly or slightly better compared with the traditional RNNs or even ConvLSTM. The procedure of computing different gates and hidden states/outputs of ConvGRU is demonstrated with equations (6)-(9), in which the symbols have the same meaning as described before, while additional $z_t$ and $r_t$ mean the update gate and the reset gate, respectively, plus $\widetilde{\mathcal{H}}$ represents the current candidate hidden representation.

$$z_t = \sigma(W_{zx} * X_t + W_{zh} * \mathcal{H}_{t-1} + b_z) \quad (6)$$
$$r_t = \sigma(W_{rx} * X_t + W_{rh} * \mathcal{H}_{t-1} + b_r) \quad (7)$$
$$\widetilde{\mathcal{H}}_t = \tanh(W_{ox} * X_t + W_{oh} * (r_t \odot \mathcal{H}_{t-1}) + b_o) \quad (8)$$
$$\mathcal{H}_t = z_t \widetilde{\mathcal{H}} + (1-z_t)\mathcal{H}_{t-1} \quad (9)$$

In ConvGRU, there are only two gate structures, i.e., the update gate $z_t$ and the reset gate $r_t$. It is the update gate $z_t$ that decides how to update the hidden representation when generating the ultimate result of $\mathcal{H}_t$ at the current layer, as shown in equation (9). While the reset gate $r_t$ is served to control to what extent the feature information captured in the previous hidden state is supposed to be forgotten through an element-wise multiplication operation when calculating current candidate hidden representation. From the equations, it is concluded that the information of $\mathcal{H}_t$ mainly comes from $\widetilde{\mathcal{H}}_t$, while $\mathcal{H}_{t-1}$ as the previous hidden-state representation also contributes to the process of computing the final representation of $\mathcal{H}_t$, thus the temporal dependencies are captured.

In practice, both ConvLSTM and ConvGRU with different numbers of hidden layers were employed to serve as the ST-RNN module in the proposed architecture, and the corresponding performances were evaluated, respectively. To be specific, in the proposed network, the input and the output sizes of the ST-RNN block are equivalent to the feature map size extracted through the encoder, which are $8 \times 16$ and $4 \times 8$ for the UNet based and SegNet based backbone, respectively. The convolutional kernel size in ConvLSTM and ConvGRU is $3 \times 3$, and the dimension of each hidden layer is 512. The detailed implementations are described in the following section.

## 2.3 Detailed implementation

*1) Network Design Details*: The proposed spatial-temporal sequence-to-one neural network was developed for the lane detection task with *K* (in this paper *K=5* if not specified) continuous image frames as inputs. The image frames were firstly fed into the encoder for feature extraction and abstraction. Different from the normal CNN-based encoder, the SCNN layer was utilized to effectively extract the spatial relationships within every image. Different locations of the SCNN layer were tested, i.e., embedding the SCNN layer after the first hidden convolutional layer or at the very beginning. The outputs of the encoder network were modelled in a time-series manner and fed into the ST-RNN blocks (i.e., ConvLSTM or ConvGRU layers) to further extract more useful and accurate features, especially the spatial-temporal dependencies and correlations among different image frames. In short, the encoder network is primarily responsible for spatial feature extraction and abstraction transforming input images into specified feature maps, while the ST-RNN blocks accept the extracted features from the continuous image frames in a time-series manner to capture the spatial-temporal dependencies.

The outputs of the ST-RNN blocks were then transferred into the decoder network that adopts deconvolution and upsampling operations to highlight and make full use of the features and rebuild the target to the original size of the input image. Note there is the skip concatenate connection (for UNet-based architecture) or pooling indices reusing (for SegNet-based architecture) between the encoder and decoder to reuse the retained features from previous encoder layers for more accurate predictions at the decoder phase. After the decoder phase, the lane detection result is obtained as an image in the equivalent size to the input image frame. With the labelled ground truth and the help of the encoder-decoder structure, the proposed model can be trained and implemented in an end-to-end way. The detailed input, output sizes, together with parameters of the layers in the entire neural network are listed in Appendix Table A1 and Table A2.

For both SegNet-based and UNet-based implementations, two types of RNN layers, i.e., ConvLSTM and ConvGRU, were tested to serve as the ST-RNN block. Besides, the ST-RNN blocks were tested with 1 hidden layer and 2 hidden layers, respectively. So there are four variants of in the proposed SegNet-based models, i.e., SCNN_SegNet_ConvGRU1, SCNN_SegNet_ConvGRU2, SCNN_SegNet_ConvLSTM1, and SCNN_SegNet_ConvLSTM2. SCNN_SegNet_ConvGRU1 means the model is using SegNet as the backbone with SCNN layer embedded encoder, and 1 hidden layer of ConvGRU as the ST-RNN block. This naming rule applies to the other 3 variants. Also, there are four variants of the proposed UNet-based models, with a similar naming rule.

In the proposed models with U-Net as the backbone, the number of kernels used in the last convolutional block of the encoder part differs from the original U-Net's settings. Here, the number of output kernels (channels) of the last convolutional block in the proposed encoder does not double its input kernels, which applies to all the previous convolutional blocks. This is done, similar to (Zou et al., 2020), to better connect the output of the encoder with the ST-RNN block (ConvLSTM or ConvGRU layers). To do so, the parameters of the full-connection layer are designed to be quadrupled while the side lengths of the feature maps reduced to half, at the same time, the number of kernels remains unchanged. This strategy also somewhat contributes to reducing the parameter size of the whole network.

A modified light version of UNet (UNetLight) was also tested to serve as the network backbone to reduce the total parameter size, increase the model's ability to operate in real-time, and also further verify the proposed network architecture's effectiveness. The UNetLight has a similar network design to the demonstration in Table A2. The only difference is that all the numbers of kernels in the ConvBlocks are reduced to half except for the *Input* in *In_ConvBlock* (with the input channel of 3 unchanged) and *Output* in *Out_ConvBlock* (with the output channel of 2 unchanged). To save space, the parameter settings of UNetLight based implementation will not be illustrated.

*2) Loss function*: Since the lane detection is modeled as a segmentation task and a pixel-wise binary classification problem, cross-entropy is a suitable candidate to serve as the loss function. However, because the pixels classified to be lanes are always quite less than those classified to be the background (meaning that it is an imbalanced binary classification and discriminative segmentation task), in the implementation, the loss was built upon the weighted cross-entropy. The adopted loss function as the standard weighted binary cross-entropy function is given as in equation (10),

$$Loss = -\frac{1}{S}\sum_{i=1}^{S}[w * y_i * log(h_\theta(x_i)) + (1-y_i) * log(1 - h_\theta(x_i))] \quad (10)$$

where $S$ is number of training examples, $w$ stands for the weight which is set according to the ratio between the total lane pixel quantities and none-lane pixel quantities throughout the whole training set, $y_i$ is the true target label for training example $i$, $x_i$ is the input for training example $i$, and $h_\theta$ stands for the model with neural network weights $\theta$.

*3) Training details*: The proposed neural networks with different variants, together with the baseline models were trained on the Dutch high-performance supercomputer clusters, Cartesius and Lisa, using 4 Titan RTX GPUs with the data parallel mechanism in PyTorch. The input image size was set as $128 \times 256$ to reduce the computational payload. The batch size was set to be as large as possible (e.g., 64 for UNet-based network architecture, 100 for SegNet based ones, and 136 for UNetLight based ones), and the learning rate was initially set to 0.03. The RAdam optimizer (Liu et al., 2019) was first used in this work for training the model at the beginning. At the later stage, when the training accuracy was beyond 95%, the optimizer was switched to the Stochastic Gradient Descent (SGD) (Bottou, 2010) optimizer with decay. With the labelled ground truth, the models were trained through iteratively updating the parameters in the weight matrices and the losses on the basis of the deviation between outputs of the proposed neural network and the ground truth using the backpropagation mechanism. To speed up the training process, the pre-trained weights of SegNet and U-Net on ImageNet (Deng et al., 2009) were adopted.

## 3 EXPERIMENTS AND RESULTS

Extensive experiments were carried out to inspect and verify the accuracy, effectiveness, and robustness of the proposed lane detection model using two large-scale open-sourced datasets. The proposed models were evaluated on different driving scenes and were compared with several state-of-the-art baseline lane detection methods which also employ deep learning, e.g., U-Net (Ronneberger et al., 2015), Seg-Net (Badrinarayanan et al., 2017), SCNN (Pan et al., 2018), LaneNet (Neven et al., 2018), UNet_ConvLSTM (Zou et al., 2020), and SegNet_ConvLSTM (Zou et al., 2020).

### 3.1 Datasets

*1) tvtLANE training set*: To verify the proposed model performance, the tvtLANE dataset (Zou et al., 2020) based upon the TuSimple lane marking challenge dataset, was first utilized for training, validating, and testing. The original dataset of the TuSimple lane marking challenge includes 3,626 clips of training and 2,782 clips of testing which are collected under various weather conditions and during different periods. In each clip, there are 20 continuous frames saved in the same folder. In each clip, only the lane marking lines of the very last frame, i.e., the 20[th] frame, are labelled with the ground truth officially. Zou et al. (2020) additionally labelled every 13[th] image in each clip and added their own collected lane dataset which includes 1,148 sequences of rural driving scenes collected in China. This immensely expanded the variety of the road and driving conditions since the original TuSimple dataset only covers the highway driving conditions. $K$ continuous frames of each clip are used as the inputs with the ground truth of the labelled $13^{th}$ or $20^{th}$ frame to train the models.

To further augment the training dataset, crop, flip, and rotation operations were employed, thus a total number of $(3{,}626 + 1{,}148) \times 4 = 19{,}096$ continuous sequences were produced, in which 38,192 images are labelled with ground truth. To adapt to different driving speeds, the input image sequences were sampled at 3 strides with a frame interval of 1, 2, or 3, respectively. Then, 3 sampling methods were employed to construct the training samples regarding the labelled $13^{th}$ and $20^{th}$ frames in each sequence, as demonstrated in Table 1.

*2) tvtLANE testing set*: Two different datasets were used for testing, i.e., Testset #1 (normal) and Testset #2 (challenging), which are also formatted with 5 continuous images as the input to detect the lane markings in the very last frame with the labelled ground truth. To be specific, Testset #1 is built upon the original TuSimple test set for normal driving scene testing; while Testset #2 is constructed with 12 challenging driving situations, especially used for robustness evaluation. The detailed descriptions of the trainset and testset in tvtLANE are illustrated in Table 1, with examples shown in Figure 2.

## 3.1 Qualitative evaluation

Qualitative evaluation with the visualization of the lane detection results is the most intuitive approach to compare and evaluate the properties of different models, and it helps to find insights regarding their pros and cons.

*1) tvtLANE Testset #1: normal situations*

Samples of the lane detection results on tvtLANE testset #1 of the proposed models and other state-of-the-art models are demonstrated in Figure 3(1). All these results are without post-processing.

In general, a good lane detection should include the following 5 properties:

• The number of lines need to be predicted correctly. A wrong detection or a misprediction might cause the automated vehicles to consider unsafe or unreachable areas as drivable areas resulting in potential accidents. As illustrated in the $1^{st}$ and $2^{nd}$ columns in Figure 3(1), the proposed models can identify the correct number of lane lines, while the baseline models, especially the ones using a single image, somewhat cannot detect the correct number of lines compared with ground truth.

• The positions of each lane marking line should be predicted precisely accords with the ground truth. As illustrated in Figure 3(1), the proposed models in row (j) with the model named by SCNN_SegNet_ConvLSTM2 and row (n) with the model named by SCNN_UNet_ConvLSTM2, could deliver better lane location predictions with thinner lines, compared with the baseline models. Superior to scattering points around, thinner predicted lane lines indicate a more precise model prediction of the lane position.

**TABLE 1.** Trainset and testset in tvtLANE.

| Trainset | | |
|---|---|---|
| Subset | | Labled Images Num |
| Original TuSimple Dataset (Highway) | | 7,252 |
| Zou et al. (2020) added (Rural Road) | | 2,296 |
| **Sample Methods** | | |
| Labled Ground Truth | Sample Stride | Train Sample Frames |
| $13^{th}$ | 3 | $1^{st}, 4^{th}, 7^{th}, 10^{th}, 13^{th}$ |
| | 2 | $5^{th}, 7^{th}, 9^{th}, 11^{th}, 13^{th}$ |
| | 1 | $9^{th}, 10^{th}, 11^{th}, 12^{th}, 13^{th}$ |
| $20^{th}$ | 3 | $8^{th}, 11^{th}, 14^{th}, 17^{th}, 20^{th}$ |
| | 2 | $12^{th}, 14^{th}, 16^{th}, 18^{th}, 20^{th}$ |
| | 1 | $16^{th}, 17^{th}, 18^{th}, 19^{th}, 20^{th}$ |
| **Testset** | | |
| Subset | Labled Images Num | Labled Ground Truth | Sample Stride | Test Sample Frames |
| Testset #1 Normal | 540 | $13^{th}$ | 1 | $9^{th}, 10^{th}, 11^{th}, 12^{th}, 13^{th}$ |
| | | $20^{th}$ | 1 | $16^{th}, 17^{th}, 18^{th}, 19^{th}, 20^{th}$ |
| Testset #2 Challenging | 728 | All | 1 | $1^{st}, 2^{nd}, 3^{rd}, 4^{th}, 5^{th}$ $2^{nd}, 3^{rd}, 4^{th}, 5^{th}, 6^{th}$ $3^{rd}, 4^{th}, 5^{th}, 6^{th}, 7^{th}$ ... |

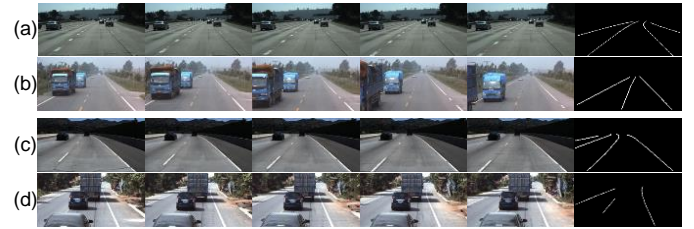

**FIGURE 2.** Samples data in trainset and testset. (a) original TuSimple dataset (Highway), (b) Zou et al., (2020) added Rural Road situations, (c) Testset #1 Normal situations, and (d) Testset #2 Challenging situations. In each row, the first five images are the input image sequence the last image is the labelled ground truth.

• The predicted lane lines should not merge or be broken. As illustrated in the $1^{st}, 2^{nd}, 6^{th}, 7^{th}$, and $8^{th}$ columns of Figure 3(1), some baseline models' output lane lines either merge at the far end or break the continuity with dashed lines. The proposed models perform slightly better although in a few cases the lines are also discontinuous.

• The lanes should be predicted correctly even at the boundary of the image. As can be found in Figure 3(1), some baseline models, e.g., row (c), (d), and (e), run across difficulties at the top boundary of the image with merge lanes on the top. This also accords with the aforementioned property.

• The lane detection models should deliver accurate predictions under different driving scenes, even under some challenging situations. For example, in the $2^{nd}, 3^{rd}, 5^{th}$, and $7^{th}$ columns of Figure 3(1), vehicles are occluding the lanes. A good lane detection model should be able to handle these. The proposed models perform well under these slightly challenging cases, more challenging situations are further discussed later.

**Input images:** (a)

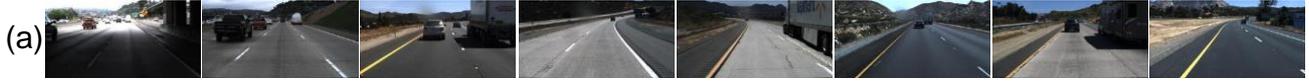

**Ground truth:** (b)

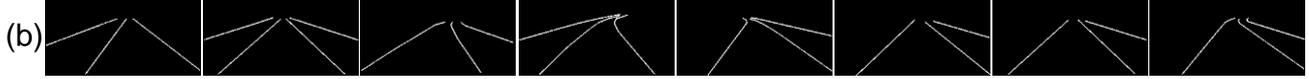

**Baseline Models:** (c) SegNet; (d) UNet; (e) SegNet_ConvLSTM; (f) UNet_ConvLSTM

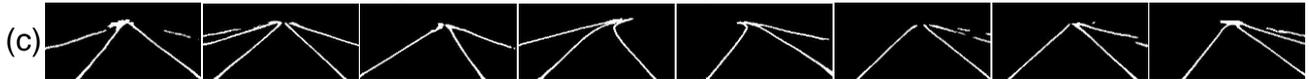
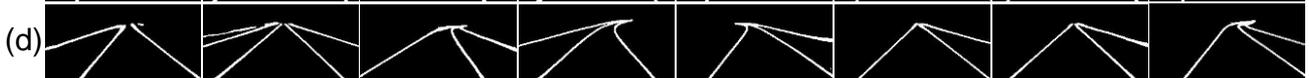
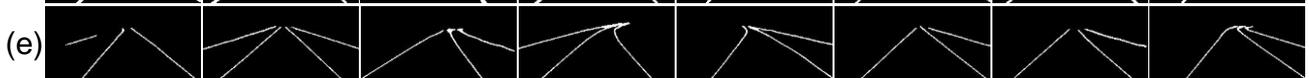
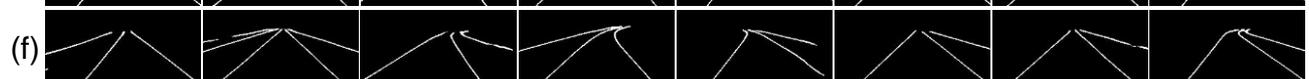

**Proposed Models SegNet-based**: (g) SCNN_SegNet_ConvGRU1; (h) SCNN_SegNet_ConvGRU2; (i) SCNN_SegNet_ConvLSTM1; (j) SCNN_SegNet_ConvLSTM2

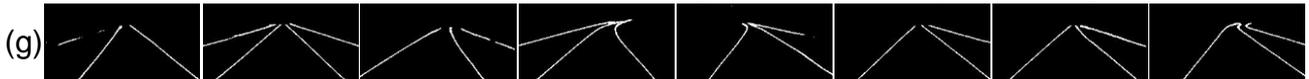
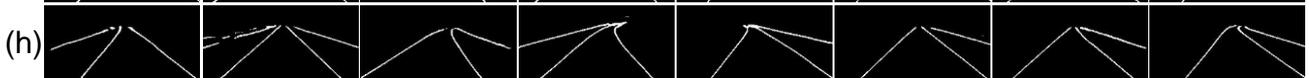
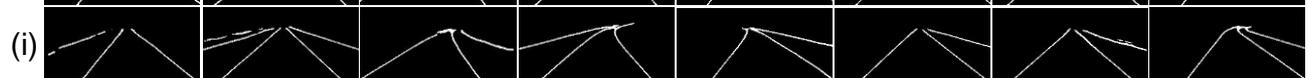
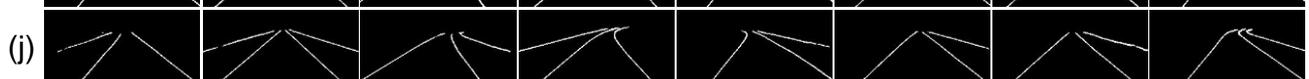

**Proposed Models UNet-based:** (k) SCNN_UNet_ConvGRU1;  (l) SCNN_UNet_ConvGRU2; (m) SCNN_UNet_ConvLSTM1; (n) SCNN_UNet_ConvLSTM2

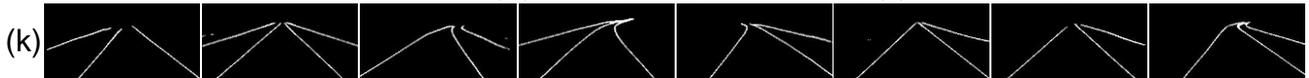
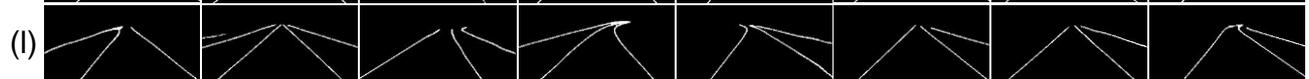
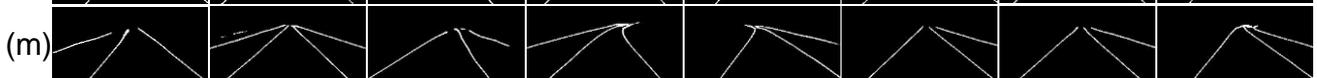
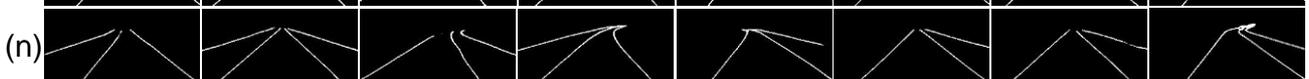

**Proposed Models UNetLight-based**: (o) SCNN_UNetLight_ConvGRU1; (p) SCNN_UNetLight_ConvGRU2; (q) SCNN_UNetLight_ConvLSTM1; (r) SCNN_UNetLight_ConvLSTM2

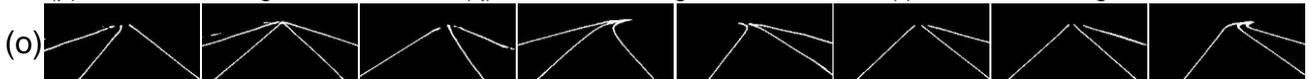
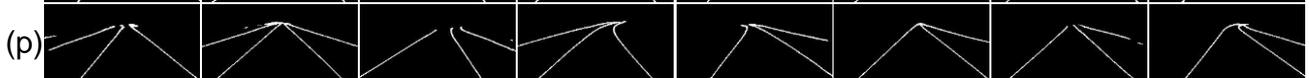
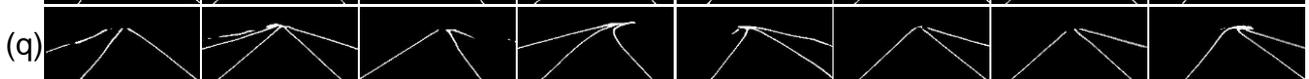
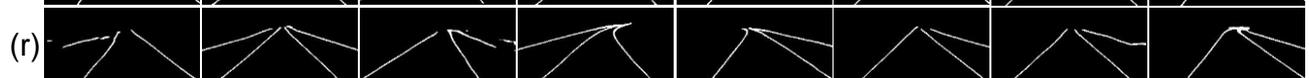

(1) Visualization of the lane-detection results on tvtLANE Testset #1 (normal situations).

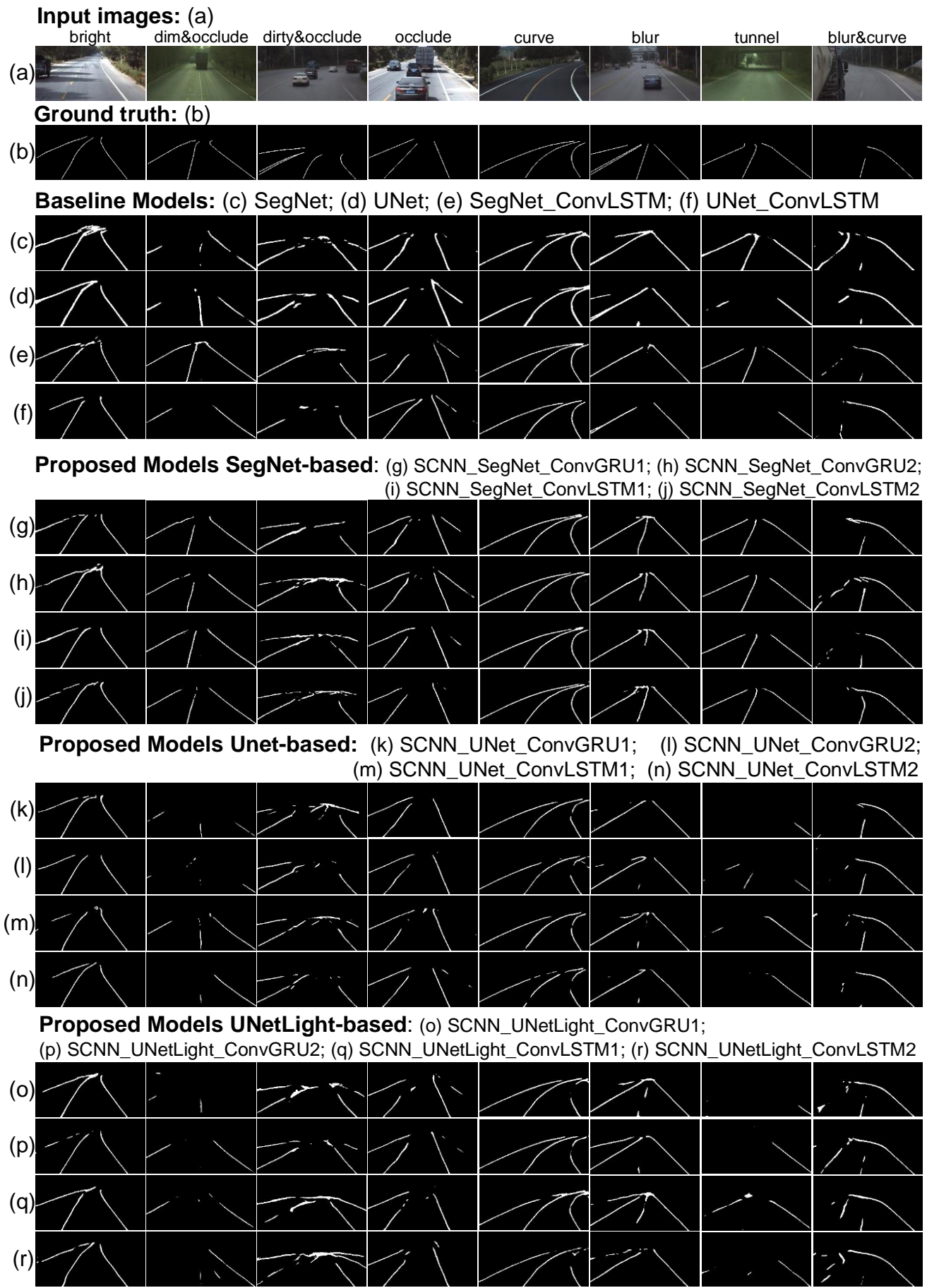

(2) Visualization of the lane-detection results on tvtLANE Testset #2 (challenging situations)

**FIGURE 3.** Qualitative evaluation: visualization of the lane-detection results on (1) tvtLANE Testset #1 and (2) tvtLANE Testset #2

*2) tvtLANE Testset #2: 12 challenging driving cases*

Figure 3(2) shows the comparison of the proposed models with the baseline models under some extremely challenging driving scenes in the tvtLANE testset#2. All the results are not post-processed. These challenging scenes cover wide situations including serious vehicle occlusion, bad lighting conditions (e.g., shadow, dim), tunnel situations, and dirt road conditions. In some extremely challenging cases, the lanes are totally occluded by vehicles, other objects, and/or shadows, which could be very difficult even for humans to do the detection.

As can be observed in Figure 3(2), although all the baseline models fail in these challenging cases, the proposed models, especially the one named SCNN_SegNet_ConvLSTM2 illustrated in the row (k), could still deliver good predictions in almost every situation listed in Figure 3(2). The only flaw is that in the 3rd column where vehicle occlusion and blur road conditions happen simultaneously, the proposed models also find it hard to predict precisely. With the results in the 4th, 7th, and 8th columns, the robustness of SCNN_SegNet_ConvLSTM2's property in detecting the correct number of lane lines is further verified, especially, one can observe in the 4th column, where almost all the other models are defeated, SCNN_SegNet_ConvLSTM2 can still predict the correct number of lanes.

Furthermore, it should be noticed that correct lane location predictions in these challenging situations are of vital importance for safe driving. For example, regarding the situation in the last column where a heavy vehicle totally shadows the field of vision on the left side, it will be very dangerous if the automated vehicle is driving according to the lane detection results demonstrated in the 3rd to 5th rows.

## 3.2 Quantitative evaluation

*1) Evaluation metrics:* This subsection examines the proposed models' properties regarding quantitative evaluations. When treated as a pixel-wise classification task, accuracy must be the most simple criterion for the performance evaluation of lane detection (Zou et al., 2017), which represents the overall classification performance in terms of correctly classified pixels, indicated in equation (11).

$$\text{Accuracy} = \frac{Truly\ Classified\ Pixels}{Total\ Number\ of\ Pixels} \quad (11)$$

However, since it is an imbalanced binary classification problem, where the lanes pixels are far less than the background pixels, using only accuracy to evaluate the model is not suitable. Thus, Precision, Recall, and F-measure, illustrated by equation (12)-(14), are commonly employed.

$$\text{Precision} = \frac{True\ Positive}{True\ Positive + False\ Positive} \quad (12)$$

$$\text{Recall} = \frac{True\ Positive}{True\ Positive + False\ Negative} \quad (13)$$

$$\text{F-measure} = (1 + \beta^2) \frac{Precision * Recall}{\beta^2 Precision + Recall} \quad (14)$$

In the above equation, true positive indicates the number of image pixels that are lane marking and are correctly identified; false positive means the number of image pixels that are background but are wrongly classified as lane markings; false negative stands for the number of image pixels which are lane marking but are wrongly classified as the background.

Specifically, this study chooses $\beta = 1$, which corresponds to the F1-measure (harmonic mean) shown in equation (15).

$$\text{F1-measure} = 2 * \frac{Precision * Recall}{Precision + Recall} \quad (15)$$

The F1-Measure, which balances Precision and Recall, is always selected as the main benchmark for model evaluation, e.g., (Liu et al., 2021; Pan et al., 2018; Xu et al., 2020; Zhang et al., 2021; Zou et al., 2020).

Furthermore, the model parameter size, i.e., Params (M), together with the multiply-accumulate (MAC) operations, i.e., MACs (G), are provided as indicators of the model complexity. The two indicators are commonly used to estimate models' computational complexities and real-time capabilities.

*2) Performance and comparisons on tvtLANE testset #1(normal situations)*

As shown in Table 2, the proposed model of SCNN_UNet_ConvLSTM2, performs the best when evaluating on tvtLANE Testset#1, with the highest Accuracy and F1-Measure, while the proposed model of SCNN_SegNet_ConvLSTM2 delivers the best Precision.

**TABLE 2. Model performance comparison on tvtLANE testset #1 (normal situations)**

| | | Test_Acc (%) | Precision | Recall | F1-Measure | MACs (G) | Params (M) |
|---|---|---|---|---|---|---|---|
| | **Baseline Models** | | | | | | |
| Models using **single image** as input | U-Net | 96.54 | 0.790 | **0.985** | 0.877 | 15.5 | 13.4 |
| | SegNet | 96.93 | 0.796 | 0.962 | 0.871 | 50.2 | 29.4 |
| | SCNN* | 96.79 | 0.654 | 0.808 | 0.722 | 77.7 | 19.2 |
| | LaneNet* | 97.94 | 0.875 | 0.927 | 0.901 | 44.5 | 19.7 |
| | SegNet_ConvLSTM** | 97.92 | 0.874 | 0.931 | 0.901 | 217.0 | 67.2 |
| | UNet_ConvLSTM** | 98.00 | 0.857 | 0.958 | 0.904 | 69.0 | 51.1 |
| | **Proposed Models (SegNet-Based)** | | | | | | |
| | SCNN_SegNet_ConvGRU1 | 98.00 | 0.878 | 0.935 | 0.905 | 219.2 | 43.7 |
| | SCNN_SegNet_ConvGRU2 | 98.05 | 0.888 | 0.918 | 0.903 | 221.5 | 57.9 |
| | SCNN_SegNet_ConvLSTM1 | 98.01 | 0.881 | 0.935 | 0.907 | 220.0 | 48.5 |
| | SCNN_SegNet_ConvLSTM2 | 98.07 | **0.893** | 0.928 | 0.910 | 223.0 | 67.3 |
| Models using **continuous images sequence** as inputs | **Proposed Models (UNet-Based)** | | | | | | |
| | SCNN_UNet_ConvGRU1 | 98.13 | 0.878 | 0.957 | 0.916 | 77.9 | 27.7 |
| | SCNN_UNet_ConvGRU2 | **98.19** | 0.887 | 0.950 | 0.917 | 87.0 | 41.9 |
| | SCNN_UNet_ConvLSTM1 | 98.18 | 0.886 | 0.948 | 0.916 | 81.0 | 32.4 |
| | SCNN_UNet_ConvLSTM2 | **98.19** | 0.889 | 0.950 | **0.918** | 93.0 | 51.3 |
| | **Proposed Models (Light Version UNet-Based)** | | | | | | |
| | SCNN_UNetLight_ConvGRU1 | 97.83 | 0.850 | 0.960 | 0.902 | 19.6 | **6.9** |
| | SCNN_UNetLight_ConvGRU2 | 98.01 | 0.863 | 0.950 | 0.905 | 21.9 | 10.5 |
| | SCNN_UNetLight_ConvLSTM1 | 97.71 | 0.830 | 0.950 | 0.886 | 20.4 | 8.1 |
| | SCNN_UNetLight_ConvLSTM2 | 97.76 | 0.840 | 0.953 | 0.893 | 23.4 | 12.8 |

* Results reported in (Zhang et al., 2021).
** There are two hidden layers of ConvLSTM in SegNet_ConvLSTM and UNet_ConvLSTM.

**TABLE 3. Model performance comparison on tvtLANE testset #2 (12 types of challenging scenes)**

| PRECISION | | | | | | | | | | | | | |
|---|---|---|---|---|---|---|---|---|---|---|---|---|---|
| Challenging Scenes / Models | 1- curve & occlude | 2- shadow- bright | 3- bright | 4- occlude | 5- curve | 6- dirty & occlude | 7- urban | 8- blur & curve | 9- blur | 10- shadow- dark | 11- tunnel | 12- dim & occlude | overall |
| U-Net | 0.7018 | 0.7441 | 0.6717 | 0.6517 | 0.7443 | 0.3994 | 0.4422 | 0.7612 | 0.8523 | 0.7881 | 0.7009 | 0.5968 | 0.6754 |
| SegNet | 0.6810 | 0.7067 | 0.5987 | 0.5132 | 0.7738 | 0.2431 | 0.3195 | 0.6642 | 0.7091 | 0.7499 | 0.6225 | 0.6463 | 0.6080 |
| UNet_ConvLSTM | 0.7591 | 0.8292 | 0.7971 | 0.6509 | 0.8845 | 0.4513 | 0.5148 | **0.8290** | **0.9484** | **0.9358** | 0.7926 | 0.8402 | **0.7784** |
| SegNet_ConvLSTM | 0.8176 | 0.8020 | 0.7200 | 0.6688 | 0.8645 | 0.5724 | 0.4861 | 0.7988 | 0.8378 | 0.8832 | 0.7733 | 0.8052 | 0.7563 |
| SCNN_SegNet_ConvGRU1 | 0.8107 | 0.7951 | 0.7225 | 0.6830 | 0.8503 | 0.4640 | 0.5071 | 0.6699 | 0.8481 | 0.8994 | 0.7804 | 0.8429 | 0.7477 |
| SCNN_SegNet_ConvGRU2 | 0.7952 | 0.8087 | 0.7770 | 0.6444 | 0.8689 | 0.5067 | 0.5171 | 0.7147 | 0.8423 | 0.8744 | 0.7979 | **0.8757** | 0.7572 |
| SCNN_SegNet_ConvLSTM1 | 0.7945 | 0.8078 | 0.7600 | 0.6417 | 0.8525 | 0.5252 | 0.3686 | 0.7582 | 0.7715 | 0.8702 | 0.7778 | 0.8517 | 0.7348 |
| SCNN_SegNet_ConvLSTM2 | 0.8326 | 0.7497 | 0.7470 | 0.7369 | 0.8647 | **0.6196** | 0.4333 | 0.7371 | 0.8566 | 0.9125 | 0.8153 | 0.8466 | 0.7673 |
| SCNN_UNet_ConvGRU1 | 0.8492 | 0.8306 | 0.8163 | **0.7845** | 0.8819 | 0.4025 | 0.4493 | 0.7378 | 0.8291 | 0.8928 | **0.8198** | 0.8040 | 0.7639 |
| SCNN_UNet_ConvGRU2 | **0.8678** | 0.7873 | **0.8548** | 0.7654 | 0.8805 | 0.5319 | 0.4735 | 0.8064 | 0.8765 | 0.8431 | 0.7112 | 0.7388 | 0.7640 |
| SCNN_UNet_ConvLSTM1 | 0.8602 | 0.7844 | 0.8119 | 0.7807 | **0.8871** | 0.4066 | 0.4652 | 0.7445 | 0.8321 | 0.8972 | 0.7507 | 0.7068 | 0.7531 |
| SCNN_UNet_ConvLSTM2 | 0.8182 | **0.8362** | 0.8189 | 0.7359 | 0.8365 | 0.5872 | **0.5377** | 0.8046 | 0.8770 | 0.8722 | 0.7952 | 0.7817 | **0.7784** |
| SCNN_UNetLight_ConvGRU1 | 0.8212 | 0.7454 | 0.7189 | 0.6996 | 0.8521 | 0.3499 | 0.3999 | 0.7851 | 0.7282 | 0.8686 | 0.6940 | 0.6289 | 0.7011 |
| SCNN_UNetLight_ConvGRU2 | 0.8147 | 0.8349 | 0.7390 | 0.7004 | 0.8591 | 0.4039 | 0.3360 | 0.6811 | 0.8300 | 0.8533 | 0.8125 | 0.7996 | 0.7238 |
| SCNN_UNetLight_ConvLSTM1 | 0.7222 | 0.7450 | 0.6533 | 0.6203 | 0.8039 | 0.2635 | 0.2716 | 0.7341 | 0.7546 | 0.7319 | 0.6298 | 0.7406 | 0.6377 |
| SCNN_UNetLight_ConvLSTM2 | 0.7618 | 0.7416 | 0.7067 | 0.6537 | 0.8096 | 0.1921 | 0.2639 | 0.6857 | 0.6830 | 0.6931 | 0.6391 | 0.6022 | 0.6190 |
| **F1-MEASURE** | | | | | | | | | | | | | |
| Challenging Scenes / Models | 1- curve & occlude | 2- shadow- bright | 3- bright | 4- occlude | 5- curve | 6- dirty & occlude | 7- urban | 8- blur & curve | 9- blur | 10- shadow- dark | 11- tunnel | 12- dim & occlude | overall |
| U-Net | 0.8200 | 0.8408 | 0.7946 | 0.7337 | 0.7827 | 0.3698 | 0.5658 | 0.8147 | 0.7715 | 0.6619 | 0.5740 | 0.4646 | 0.6985 |
| SegNet | 0.8042 | 0.7900 | 0.7023 | 0.6127 | 0.8639 | 0.2110 | 0.4267 | 0.7396 | 0.7286 | 0.7675 | 0.6935 | 0.5822 | 0.6727 |
| UNet_ConvLSTM | 0.8465 | **0.8891** | **0.8411** | 0.7245 | 0.8662 | 0.2417 | 0.5682 | 0.8323 | 0.7852 | 0.6404 | 0.4741 | 0.5718 | 0.7143 |
| SegNet_ConvLSTM | 0.8852 | 0.8544 | 0.7688 | 0.6878 | 0.9069 | 0.4128 | 0.5317 | 0.7873 | 0.7575 | 0.8503 | 0.7865 | 0.7947 | 0.7609 |
| SCNN_SegNet_ConvGRU1 | 0.8821 | 0.8626 | 0.7734 | 0.7185 | 0.9039 | 0.3027 | 0.5288 | 0.7229 | **0.7866** | 0.8658 | 0.7759 | 0.7763 | 0.7547 |
| SCNN_SegNet_ConvGRU2 | 0.8710 | 0.8630 | 0.8094 | 0.6989 | 0.9005 | 0.3963 | 0.5497 | 0.7470 | 0.7637 | 0.8525 | 0.7798 | 0.7396 | 0.7591 |
| SCNN_SegNet_ConvLSTM1 | 0.8768 | 0.8801 | 0.8185 | 0.7166 | 0.9083 | 0.3750 | 0.4516 | 0.7806 | 0.7320 | 0.8622 | **0.8029** | 0.8245 | 0.7629 |
| SCNN_SegNet_ConvLSTM2 | 0.8956 | 0.8237 | 0.7909 | 0.7468 | **0.9108** | 0.4398 | 0.4858 | 0.7379 | 0.7546 | **0.8729** | 0.7963 | 0.8074 | **0.7666** |
| SCNN_UNet_ConvGRU1 | 0.8608 | 0.8745 | 0.8393 | **0.7802** | 0.9005 | 0.3181 | 0.5143 | 0.7833 | 0.7567 | 0.5554 | 0.3503 | 0.3703 | 0.6839 |
| SCNN_UNet_ConvGRU2 | 0.8706 | 0.8556 | 0.8304 | 0.7647 | 0.8532 | 0.3515 | 0.5253 | **0.8345** | 0.7399 | 0.5405 | 0.3567 | 0.2855 | 0.6722 |
| SCNN_UNet_ConvLSTM1 | **0.8971** | 0.8493 | 0.8234 | 0.7633 | 0.8997 | 0.3054 | 0.5307 | 0.7424 | 0.7436 | 0.6243 | 0.5568 | 0.5366 | 0.6992 |
| SCNN_UNet_ConvLSTM2 | 0.8670 | 0.8866 | 0.8405 | 0.7565 | 0.7955 | 0.4179 | **0.5933** | 0.7880 | 0.7285 | 0.6296 | 0.4747 | 0.4134 | 0.7024 |
| SCNN_UNetLight_ConvGRU1 | 0.8896 | 0.8212 | 0.7819 | 0.7517 | 0.8913 | 0.3043 | 0.4961 | 0.8133 | 0.7000 | 0.5635 | 0.3086 | 0.2733 | 0.6637 |
| SCNN_UNetLight_ConvGRU2 | 0.8593 | 0.8730 | 0.7878 | 0.7406 | 0.8889 | 0.3335 | 0.4266 | 0.7263 | 0.7782 | 0.6498 | 0.5280 | 0.5257 | 0.6910 |
| SCNN_UNetLight_ConvLSTM1 | 0.8115 | 0.8056 | 0.7168 | 0.6882 | 0.8179 | 0.2613 | 0.3681 | 0.7834 | 0.7576 | 0.5701 | 0.5281 | 0.5081 | 0.6418 |
| SCNN_UNetLight_ConvLSTM2 | 0.8377 | 0.8158 | 0.7620 | 0.6971 | 0.8365 | 0.2209 | 0.3577 | 0.7551 | 0.6594 | 0.4597 | 0.3545 | 0.3559 | 0.6079 |

Incorporating the quantitative evaluation with the qualitative evaluation, it could be easily interpreted that the highest Precision, Accuracy, and F1-Measure are mainly derived from (i) the correct lane number, (ii) the accurate lane position, (iii) the sound continuity in the detected lanes, and (iv) the thinness of the predicted lanes with less blurriness, which accords with (ii). The correct prediction directly reduces the number of False Positives, and a good Precision contributes to better Accuracy and F1-Measure. Considering the structure of the proposed model architecture, a further explanation of the high F1-Measure, Accuracy, and Precision can be explained as follows:

Firstly, the SCNN layer embedded in the encoder equips the proposed model with better information extracting ability regarding the low-level features and spatial relations in each image.

Secondly, the ST-RNN blocks, i.e., ConvLSTM / ConvGRU layers, can effectively capture the temporal dependencies among the continuous image frames, which could be very helpful for challenging situations where the lanes are shadowed or covered by other objects in the current frame.

Finally, the proposed architecture could make the best of the spatial-temporal information among the processed $K$ continuous frames by regulating the weights of the convolutional kernels within the SCNN and ConvLSTM / ConvGRU layers.

All in all, with the proposed architecture the proposed model tries to not only strengthen feature extraction regarding spatial relation in one image frame but also the spatial-temporal correlation and dependencies among image frames for lane detection.

Looking at the main metric, F1-Measure, it is demonstrated that increasing only Precision or only Recall will not improve the F1-Measure. Although the bassline models of U-Net, SegNet, and SegNet_ConvLSTM get better Recalls, they do not deliver good F1-Measure since their Precisions is much lower than the proposed model of SCNN_SegNet_ConvLSTM2 or SCNN_UNet_ConvLSTM2. Regarding the good Recall of U-Net and SegNet, it could be speculated from the qualitative evaluation, where one can find that U-Net and SegNet tend to produce thicker lane lines. With thicker lines and blurry areas, the two models can somehow reduce the False Negative, which will contribute to better Recall. This also demonstrates that Recall and Precision antagonize each other which further proves that F1-Measure should be a more reasonable evaluation measure compared with Precision and Recall.

*3) Performance and comparisons on tvtLANE testset #2 (challenging situations)*

To further evaluate the proposed models' performance and verify the models' robustness, the models were evaluated on a brand-new dataset, i.e., the tvtLANE Testset #2. As introduced in *3.1 Datasets*, tvtLANE Testset #2 includes 728 images in highway, urban, and rural driving scenes. These challenging driving scenes' data were obtained by data recorders at various locations, outside and inside the car front windshield under different road and weather conditions. Testset #2 is a challenging and comprehensive dataset for model evaluation, from which some cases would be difficult enough for humans to do the correct detection.

Table 3 demonstrates the model performance comparison on the 12 types of challenging scenes in tvtLANE Testset #2. Following the results and discussions in 2) Performance and comparisons on tvtLANE testset #1(normal situations), here Table 3 provides the Precision and F1-Measure for the evaluation reference.

As indicated by the bold numbers, the proposed model, SCNN_SegNet_ConvLSTM2, results in the best F1-Measure at the overall level and in more situations, while the UNet_ConvLSTM results in the best Precision at the overall level and in more situations. Incorporating with the qualitative evaluation in Figure 3(2), it is shown that UNet_ConvLSTM tends to not classify pixels into lane lines for uncertain areas under some challenging situations (e.g., the 2$^{nd}$ and 7$^{th}$ columns in Figure 3(2)). This might be the reason for its obtaining better Precision. To further confirm this speculation, Figure 4 compares the lane detection results of SCNN_SegNet_ConvLSTM2 and UNet_ConvLSTM under challenging situations **8-blur&curve**, and **10-shadow-dark**, where UNet_ConvLSTM delivers very good Precisions.

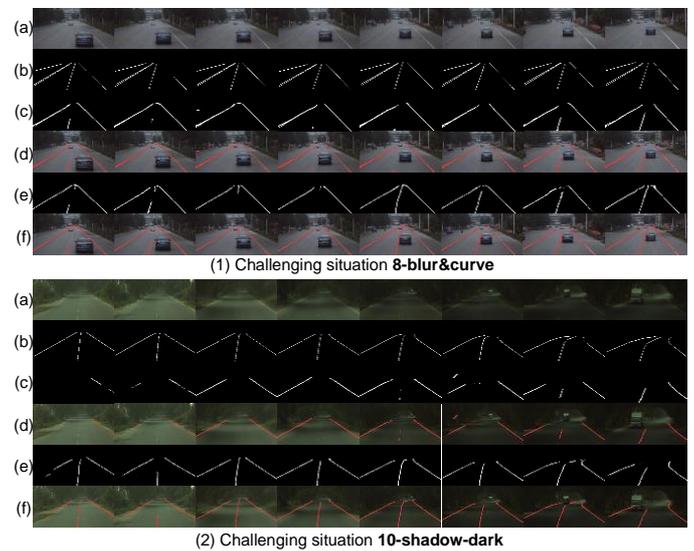

**FIGURE 4.** Visual comparison of the lane-detection results on challenging driving situations for UNet_ConvLSTM and the proposed model SCNN_SegNet_ConvLSTM2. All the results are not post-processed. (a) Input images. (b) Ground truth. (c) Detection results of UNet_ConvLSTM. (d) Detection results of UNet_ConvLSTM overlapping on the original images. (e) Detection results of SCNN_SegNet_ConvLSTM2. (f) Detection results of SCNN_SegNet_ConvLSTM2 overlapping on the original images. The upper part (1) is for challenging situation **8-blur&curve**, while the down part (2) is for situation **10-shadow-dark**.

As illustrated in Figure 4, truly UNet_ConvLSTM tries not to classify pixels into lane lines under uncertain areas as much as possible. This leads to fewer False Negatives which helps for raising a better Precision. However, in real application scenarios, this is not wise and not acceptable. On the contrary, the proposed model SCNN_SegNet_ConvLSTM2 tries to make tough but valuable detections classifying candidate points into lane lines in the challenging uncertain areas with dirt, dark road conditions, and/or vehicle occlusions. This may lead to more False Negatives and a worse Precision but is praiseworthy. These analyses further demonstrate that F1-Measure is a better measure compared with Precision. Finally, it can be concluded that the proposed model, SCNN_SegNet_ConvLSTM2, delivers the best performance on the challenging tvtLANE Testset #2, which verified the proposed model architecture's robustness.

To sum up, the proposed model architecture demonstrates its effectiveness in both normal and challenging driving scenes, with the UNet based model, SCNN_UNet_ConvLSTM2, beats the baseline models with a large margin on normal situations, while the SegNet based model, SCNN_SegNet_ConvLSTM2 performs the best handling almost all the challenging driving scenes. The finding that, compared with UNet based models, SegNet based neural network models are more robust coping with challenging driving environments accords with results in (Zou et al., 2020).

## 3.3 Parameter analysis and ablation study

*1) The added value of SCNN*

Regarding the neural network architecture, the effects of SCNN were investigated by evaluating performances of the model variants with and without SCNN layers. As demonstrated in Figure 3 and Figure 4, together with the quantitative results in Table 2 and Table 3, the proposed SegNet and UNet based models with SCNN embedded encoder, i.e., SCNN_SegNet_ConvLSTM, SCNN_SegNet_ConvGRU, SCNN_UNet_ConvLSTM, and SCNN_UNet_ConvGRU, outperform SegNet_ConvLSTM and UNet_ConvLSTM which are also SegNet or UNet based sequential model using multiple continuous image frames as inputs but without SCNN. Especially, SCNN_UNet_ConvLSTM2 obtains the best result in normal testing while SCNN_SegNet_ConvLSTM2 delivers the best performance in challenging situations.

For normal cases' testing on tvtLANE Testset#1, as shown in Table 2, by adding SCNN layer in the encoder, almost all the proposed models with SCNN embedded encoder outperform the baseline models with better F1-Measure. To be specific, SCNN_SegNet_ConvLSTM2 improves the lane detection accuracy by around 0.3% and F1-measure by around 1%, and these improvements are from the already very good results obtained by SegNet_ConvLSTM. Similarly, SCNN_UNet_ConvLSTM2 overperforms UNet_ConvLSTM with even larger margins regarding both Accuracy, Precision, and F1-measure.

For challenging situations, adding the SCNN layer also helps the proposed model, SCNN_SegNet_ConvLSTM2, beat other baseline models, and deliver the best F1-Measure as indicated in Table 3.

Figure 5 visualizes the extracted features at Down_ConvBlock_1 layer for UNet based models, with and without SCNN. Clearly, vast differences can be witnessed between the baseline model UNet_ConvLSTM and the proposed model SCNN_UNet_ConvLSTM2. In Figure 5 (b), the CNN-based UNet layers identify the low-level features in the images regarding the target lane lines. However, the extracted features are not so clear, i.e., there are some interference signals, especially as visualized in the third image of row (b), which is supposed to affect the model training (i.e., updating weight parameters of the neural networks) and thus affect the model's performance regarding the marking detection results. It might further influence the final detection results. In contrast, with SCNN layers, the extracted features of the lanes are more inerratic, clear, and evident as shown in Figure 5 (c). There are fewer interferences surrounding the detected lane features. This verifies SCNN's powerful strength in detecting the spatial relations in every single image with its message passing mechanism.

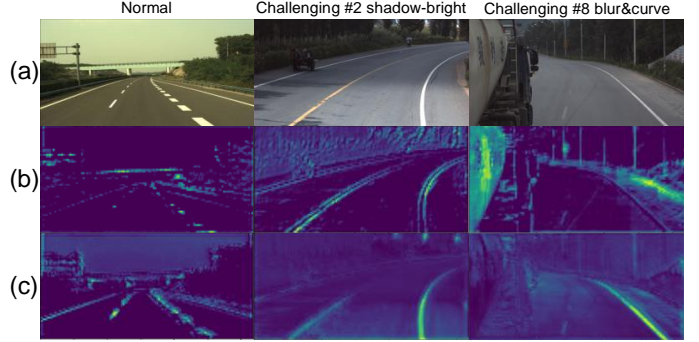

**FIGURE 5.** Visualization of the extracted low-level features at Down_ConvBlock_1 for UNet based models. (a) Original image. (b) Results of UNet_ConvLSTM (without SCNN layers). (c) Results of the SCNN_UNet_ConvLSTM2 (with SCNN layers).

All the above results demonstrate the adding of the SCNN layer embedded in the encoder does contribute to the spatial feature extraction, with which the model could better make the utmost use of the spatial-temporal information among the continuous image frames.

*2) Different locations of SCNN layer*

**TABLE 4. Model performance comparison with different locations of SCNN layer on tvtLANE testset #1 and #2.**

| Models | Location of SCNN | Testset #1 (Normal Situations) | | | | Testset #2 (Challenging Scenes) | | | |
|---|---|---|---|---|---|---|---|---|---|
| | | Test_Acc (%) | Precision | Recall | F1-Measure | Test_Acc (%) | Precision | Recall | F1-Measure |
| SegNet_ConvLSTM | Without | 97.92 | 0.874 | 0.931 | 0.901 | 97.83 | 0.756 | 0.765 | 0.761 |
| SCNN_SegNet_ConvLSTM2 | Conv1_1 | 98.00 | 0.884 | 0.921 | 0.902 | 97.92 | 0.757 | 0.757 | 0.757 |
| | Conv2_1 | 98.07 | 0.893 | 0.928 | 0.910 | 97.90 | 0.767 | 0.766 | 0.767 |
| UNet_ConvLSTM | Without | 98.00 | 0.857 | 0.957 | 0.904 | 97.93 | 0.778 | 0.660 | 0.714 |
| SCNN_UNet_ConvLSTM2 | In_Conv_1 | 98.28 | 0.896 | 0.939 | 0.917 | 98.08 | 0.776 | 0.593 | 0.672 |
| | Conv1_1 | 98.19 | 0.889 | 0.950 | 0.918 | 97.95 | 0.778 | 0.640 | 0.702 |

Results of testing different locations of the SCNN layer in the proposed model architecture are shown in Table 4. The results reveal that: (a) Compared with baseline models without SCNN layers, the embedding of SCNN layers really help to improve the models' performance, which further verifies the added-value of SCNN and accords with the aforementioned results in *1)*; (b) In terms of the main evaluation metric F1-measure, embedding SCNN layer after the Conv1_1 (in SegNet based model) or In_Conv_1 (in UNet based model) layer delivers better results compared with embedding it at the very beginning or early layers of the encoder; (c) For UNet based model, embedding SCNN layer at the very beginning delivers quite good Precision and Accuracy, but worse Recall, which means there are fewer False Positives but more False Negatives. This should be related to the properties of the UNet style neural network. These results further confirm the effectiveness of the proposed model architecture.

*3) Type and number of ST-RNN layers*

As described in Section 3, in the proposed model architecture two types of RNNs, i.e., ConvLSTM and ConvGRU, are employed to serve in the ST-RNN block, to capture and make use of the spatial-temporal dependencies and correlations among the continuous image sequences. The number of hidden ConvLSTM and ConvGRU layers were also tested from 1 to 2. The quantitative results are demonstrated in Table 2 and Table 3, while some intuitive qualitative insights could be drawn from Figure 3 and Figure 4.

From Table 2, it is illustrated that in general models adopting ConvLSTM layers in the ST-RNN block perform better than those adopting ConvGRU layers with improved F1-measure, except for the UNetLight based. This could be explained by ConvLSTM's better properties in extracting spatial-temporal features and capturing time dependencies by more control gates and thus more parameters compared with ConvGRU. Furthermore, from Table 2 and Table 3, it is observed that models with two hidden ST-RNN layers, for both ConvLSTM and ConvGRU, generally perform better than those with only one hidden ST-RNN layer. This could be speculated that with two hidden ST-RNN layers, one layer can serve for sequential feature extraction, and the other can achieve spatial-temporal feature integration. The improvements of two ST-RNN layers over one are not that significant which might be due to (a) models employing one ST-RNN layer already obtain good results; (b) since the length of the continuous image frames is only five, one ST-RNN layer might be already enough to do the spatial-temporal feature extraction, so when incorporating longer image sequences the superiorities of two ST-RNN layers could be promoted. However, longer image sequences require more computational resources and longer training time, which could not be afforded at the present stage in this study. This could be the future research direction.

*4) Number of parameters and real-time capability*

As shown in Table 2, the two proposed candidate models, i.e., SCNN_SegNet_ConvLSTM2 and SCNN_UNet_ConvLSTM2, possess a bit more parameters compared with the baseline SegNet_ConvLSTM and UNet_ConvLSTM, respectively. However, almost all of the proposed model variants with different types and numbers of ST-RNN layers outperform the baselines, and some of them are even with low parameter sizes e.g., SCNN_SegNet_ConvGRU1, SCNN_SegNet_ConvLSTM1, SCNN_UNet_ConvGRU1, SCNN_UNet_ConvLSTM1. Generally speaking, lower numbers of model parameters mean better real-time capability.

In addition, four model variants were implemented with a modified light version of UNet, i.e., UNetLight, serving as the network backbone to reduce the total parameter size and improve the model's ability to operate in real-time. The UNetLight backbone has a similar network design with UNet whose parameter settings are demonstrated in Table A2. The only difference is that all the numbers of kernels in the ConvBlocks are reduced to half except for the Input in *In_ConvBlock* (with the input channel of 3 unchanged) and Output in *Out_ConvBlock* (with the output channel of 2 unchanged). From the testing results in Table 2, it is shown that the model named SCNN_UNetLight_ConvGRU2, with fewer parameters than all the baseline models, beat the baselines exhibiting better performance regarding both Accuracy and F1-Measure. To be specific, compared with the best baseline model, i.e., UNet_ConvLSTM, SCNN_UNetLight_ConvGRU2 only uses less than one-fifth of the parameter size but delivers better evaluation metrics in testing Accuracy, Precision, and F1-Measure.

Regarding UNetLight based models, models using ConvGRU layers in the ST-RNN block perform better than those adopting ConvLSTM. The reason could be that light version UNet cannot implement high-quality feature extraction which does not feed enough information for ConvLSTM, while ConvGRU, with fewer control gates, is more robust when low-level features are not that fully extracted.

All these results further verify the proposed network architecture's effectiveness and strength.

## 4 CONCLUSION

In this paper, a novel spatial-temporal sequence-to-one model framework with a hybrid neural network architecture is proposed for robust lane detection under various normal and challenging driving scenes. This architecture integrates single image feature extraction module with SCNN, spatial-temporal feature integration module with ST-RNN, together with the encoder-decoder structure. The proposed architecture achieved significantly better results in comparison to baseline models that use a single frame (e.g., U-Net, SegNet, and LaneNet), as well as the state-of-art models adopting "CNN+RNN" structures (e.g., UNet_ConvLSTM, SegNet_ConvLSTM), with the best testing Accuracy, Precision, F1-measure on the normal driving dataset (i.e., tvtLANE Testset #1) and the best F1-measure on 12 challenging driving scenarios dataset (tvtLANE Testset #2). The results demonstrate the effectiveness of strengthening spatial relation abstraction in every single image with SCNN layer, plus the employment of multiple continuous image sequences as inputs. The results also demonstrate the proposed model architecture's ability in making the best of the spatial-temporal information in continuous image frames. Extensive experimental results show the superiorities of the sequence-to-one "SCNN + ConvLSTM" over "SCNN + ConvGRU" and ordinary "CNN + ConvLSTM" regarding sequential spatial-temporal feature extracting and learning, together with target-information classification for robust lane detection. In addition, testing results of the model variants with the modified light version of UNet (i.e., UNetLight) as the backbone, demonstrate the proposed model architecture's potential regarding real-time capability.

To the best of the authors' knowledge, the proposed model is the first attempt that tries to strengthen both spatial relations regarding feature extraction in every image frame together with the spatial-temporal correlations and dependencies among image frames for lane detection, and the extensive evaluation experiments demonstrate the strength of this proposed architecture. Therefore, it is recommended in future research to incorporate both aspects to obtain better performance.

In this paper, the challenging cases do not include night driving, rainy or wet road conditions, neither do they include situations in which the input images are defective (e.g., partly masked or blurred). There are demands to build larger test sets with comprehensive challenging situations to further validate the model's robustness. Since a large amount of unlabeled driving scene data involving various challenging cases was collected within the research group, a future research direction might be to develop semi-supervised learning methods and employ domain adaption to label the collected data, and then open source them for boosting the research in the field of robust lane detection. Furthermore, to further enhance the lane detection model, customed loss function, pre-trained techniques adopted in image-inpainting task, e.g., masked autoencoders, plus sequential attention mechanism could be introduced and integrated into the proposed framework.

## ACKNOWLEDGMENT

This work was supported by the Applied and Technical Sciences (TTW), a subdomain of the Dutch Institute for Scientific Research (NWO) through the Project Safe and Efficient Operation of Automated and Human-Driven Vehicles in Mixed Traffic (SAMEN) under Contract 17187. The authors thank Dr. Qin Zou, Hanwen Jiang, and Qiyu Dai from Wuhan University, as well as Jiyong Zhang from Southwest Jiaotong University for their tips in using the tvtLANE dataset.

# APPENDIX

See Table A1 and Table A2.

**TABLE A1. Parameter settings for each layer of the SegNet-based neural network.**

| Layer | | Input (channel×hight×width) | Output (channel×hight×width) | Kernel | Padding | Stride | Activation |
|---|---|---|---|---|---|---|---|
| Down_ConvBlock_1 | Conv_1_1 | 3×128×256 | 64×128×256 | 3×3 | (1,1) | 1 | ReLU |
| | Conv_1_2 | 64×128×256 | 64×128×256 | 3×3 | (1,1) | 1 | ReLU |
| | Maxpool1 | 64×128×256 | 64×64×128 | 2×2 | (0,0) | 2 | --- |
| SCNN | SCNN_Down | 64×1×128 | 64×1×128 | 1×9 | (0,4) | 1 | ReLU |
| | SCNN_Up | 64×1×128 | 64×1×128 | 1×9 | (0,4) | 1 | ReLU |
| | SCNN_Right | 64×64×1 | 64×64×1 | 9×1 | (4,0) | 1 | ReLU |
| | SCNN_Left | 64×64×1 | 64×64×1 | 9×1 | (4,0) | 1 | ReLU |
| Down_ConvBlock_2 | Conv_2_1 | 64×64×128 | 128×64×128 | 3×3 | (1,1) | 1 | ReLU |
| | Conv_2_2 | 128×64×128 | 128×64×128 | 3×3 | (1,1) | 1 | ReLU |
| | Maxpool2 | 128×64×128 | 128×32×64 | 2×2 | (0,0) | 2 | --- |
| Down_ConvBlock_3 | Conv_3_1 | 128×32×64 | 256×32×64 | 3×3 | (1,1) | 1 | ReLU |
| | Conv_3_2 | 256×32×64 | 256×32×64 | 3×3 | (1,1) | 1 | ReLU |
| | Conv_3_3 | 256×32×64 | 256×32×64 | 3×3 | (1,1) | 1 | ReLU |
| | Maxpool3 | 256×64×128 | 256×16×32 | 2×2 | (0,0) | 2 | --- |
| Down_ConvBlock_4 | Conv_4_1 | 256×16×32 | 512×16×32 | 3×3 | (1,1) | 1 | ReLU |
| | Conv_4_2 | 512×16×32 | 512×16×32 | 3×3 | (1,1) | 1 | ReLU |
| | Conv_4_3 | 512×16×32 | 512×16×32 | 3×3 | (1,1) | 1 | ReLU |
| | Maxpool4 | 512×16×32 | 512×8×16 | 2×2 | (0,0) | 2 | --- |
| Down_ConvBlock_5 | Conv_5_1 | 512×8×16 | 512×8×16 | 3×3 | (1,1) | 1 | ReLU |
| | Conv_5_2 | 512×8×16 | 512×8×16 | 3×3 | (1,1) | 1 | ReLU |
| | Conv_5_3 | 512×8×16 | 512×8×16 | 3×3 | (1,1) | 1 | ReLU |
| | Maxpool5 | 512×8×16 | 512×4×8 | 2×2 | (0,0) | 2 | --- |
| ST-RNN Layer1* | 5 * ConvLSTMCell(input=(512×4×8), kernel=(3,3), stride=(1,1), padding=(1,1)) **Or** 5 * ConvGRUCell(input=(512×4×8), kernel=(3,3), stride=(1,1), padding=(1,1), dropout(0.5)) | | | | | | |
| ST-RNN Layer2** | 5 * ConvLSTMCell(input=(512×4×8), kernel=(3,3), stride=(1,1), padding=(1,1)) **Or** 5 * ConvGRUCell(input=(512×4×8), kernel=(3,3), stride=(1,1), padding=(1,1), dropout(0.5)) | | | | | | |
| Up_ConvBlock_5 | MaxUnpool1 | 512×4×8 | 512×8×16 | 2×2 | (0,0) | 2 | --- |
| | Up_Conv_5_1 | 512×8×16 | 512×8×16 | 3×3 | (1,1) | 1 | ReLU |
| | Up_Conv_5_2 | 512×8×16 | 512×8×16 | 3×3 | (1,1) | 1 | ReLU |
| | Up_Conv_5_3 | 512×8×16 | 512×8×16 | 3×3 | (1,1) | 1 | ReLU |
| Up_ConvBlock_4 | MaxUnpool2 | 512×8×16 | 512×16×32 | 2×2 | (0,0) | 2 | --- |
| | Up_Conv_4_1 | 512×16×32 | 512×16×32 | 3×3 | (1,1) | 1 | ReLU |
| | Up_Conv_4_2 | 512×16×32 | 512×16×32 | 3×3 | (1,1) | 1 | ReLU |
| | Up_Conv_4_3 | 512×16×32 | 256×16×32 | 3×3 | (1,1) | 1 | ReLU |
| Up_ConvBlock_3 | MaxUnpool3 | 256×16×32 | 256×32×64 | 2×2 | (0,0) | 2 | --- |
| | Up_Conv_3_1 | 256×32×64 | 256×32×64 | 3×3 | (1,1) | 1 | ReLU |
| | Up_Conv_3_2 | 256×32×64 | 256×32×64 | 3×3 | (1,1) | 1 | ReLU |
| | Up_Conv_3_3 | 256×32×64 | 128×32×64 | 3×3 | (1,1) | 1 | ReLU |
| Up_ConvBlock_2 | MaxUnpool4 | 128×32×64 | 128×64×128 | 2×2 | (0,0) | 2 | --- |
| | Up_Conv_2_1 | 128×64×128 | 128×64×128 | 3×3 | (1,1) | 1 | ReLU |
| | Up_Conv_2_2 | 128×64×128 | 64×64×128 | 3×3 | (1,1) | 1 | ReLU |
| Up_ConvBlock_1 | MaxUnpool5 | 64×64×128 | 64×128×256 | 2×2 | (0,0) | 2 | --- |
| | Up_Conv_1_1 | 64×128×256 | 64×128×256 | 3×3 | (1,1) | 1 | ReLU |
| | Up_Conv_1_2 | 64×128×256 | 2×128×256 | 3×3 | (1,1) | 1 | LogSoftmax |

*Abbreviations*: ConvGRU, convolutional gated recurrent unit; ConvLSTM, convolutional long short-term memory; SCNN, spatial convolutional neural network; ST-RNN, spatial-temporal recurrent neural network; ReLU, Rectified Linear Unit.

* Two types of ST-RNN, i.e., ConvLSTM and ConvGRU are tested;
** ST-RNN blocks are tested with 1 hidden layer or 2 hidden layers.

**TABLE A2. Parameter settings for each layer of the UNet-based neural network.**

| Layer | | Input (channel×hight×width) | Output (channel×hight×width) | Kernel | Padding | Stride | Activation |
|---|---|---|---|---|---|---|---|
| In_ConvBlock | In_Conv_1 | 3×128×256 | 64×128×256 | 3×3 | (1,1) | 1 | ReLU |
| | In_Conv_2 | 64×128×256 | 64×128×256 | 3×3 | (1,1) | 1 | ReLU |
| SCNN | SCNN_Down | 64×1×256 | 64×1×256 | 1×9 | (0,4) | 1 | ReLU |
| | SCNN_Up | 64×1×256 | 64×1×256 | 1×9 | (0,4) | 1 | ReLU |
| | SCNN_Right | 64×128×1 | 64×128×1 | 9×1 | (4,0) | 1 | ReLU |
| | SCNN_Left | 64×128×1 | 64×128×1 | 9×1 | (4,0) | 1 | ReLU |
| Down_ConvBlock_1 | Maxpool1 | 64×128×256 | 64×64×128 | 2×2 | (0,0) | 2 | --- |
| | Conv_1_1 | 64×64×128 | 128×64×128 | 3×3 | (1,1) | 1 | ReLU |
| | Conv_1_2 | 128×64×128 | 128×64×128 | 3×3 | (1,1) | 1 | ReLU |
| Down_ConvBlock_2 | Maxpool2 | 128×64×128 | 128×32×64 | 2×2 | (0,0) | 2 | --- |
| | Conv_2_1 | 128×32×64 | 256×32×64 | 3×3 | (1,1) | 1 | ReLU |
| | Conv_2_2 | 256×32×64 | 256×32×64 | 3×3 | (1,1) | 1 | ReLU |
| Down_ConvBlock_3 | Maxpool3 | 256×32×64 | 256×16×32 | 2×2 | (0,0) | 2 | --- |
| | Conv_3_1 | 256×16×32 | 512×16×32 | 3×3 | (1,1) | 1 | ReLU |
| | Conv_3_2 | 512×16×32 | 512×16×32 | 3×3 | (1,1) | 1 | ReLU |
| Down_ConvBlock_4 | Maxpool4 | 512×16×32 | 512×8×16 | 2×2 | (0,0) | 2 | --- |
| | Conv_4_1 | 512×8×16 | 512×8×16 | 3×3 | (1,1) | 1 | ReLU |
| | Conv_4_2 | 512×8×16 | 512×8×16 | 3×3 | (1,1) | 1 | ReLU |
| ST-RNN Layer1* | 5 * ConvLSTMCell(input=(512×8×16), kernel=(3,3), stride=(1,1), padding=(1,1)) *Or* 5 * ConvGRUCell(input=(512×8×16), kernel=(3,3), stride=(1,1), padding=(1,1), dropout(0.5)) | | | | | | |
| ST-RNN Layer2** | 5 * ConvLSTMCell(input=(512×8×16), kernel=(3,3), stride=(1,1), padding=(1,1)) *Or* 5 * ConvGRUCell(input=(512×8×16), kernel=(3,3), stride=(1,1), padding=(1,1), dropout(0.5)) | | | | | | |
| Up_ConvBlock_4 | UpsamplingBilinear2D_1 | 512×8×16 | 512×16×32 | 2×2 | (0,0) | 2 | --- |
| | Up_Conv_4_1 | 1024×16×32 | 256×16×32 | 3×3 | (1,1) | 1 | ReLU |
| | Up_Conv_4_2 | 256×16×32 | 256×16×32 | 3×3 | (1,1) | 1 | ReLU |
| Up_ConvBlock_3 | UpsamplingBilinear2D_2 | 256×16×32 | 256×32×64 | 2×2 | (0,0) | 2 | --- |
| | Up_Conv_3_1 | 512×32×64 | 128×32×64 | 3×3 | (1,1) | 1 | ReLU |
| | Up_Conv_3_2 | 128×32×64 | 128×32×64 | 3×3 | (1,1) | 1 | ReLU |
| Up_ConvBlock_2 | UpsamplingBilinear2D_3 | 128×32×64 | 128×64×128 | 2×2 | (0,0) | 2 | --- |
| | Up_Conv_2_1 | 256×64×128 | 64×64×128 | 3×3 | (1,1) | 1 | ReLU |
| | Up_Conv_2_2 | 64×64×128 | 64×64×128 | 3×3 | (1,1) | 1 | ReLU |
| Up_ConvBlock_1 | UpsamplingBilinear2D_4 | 64×64×128 | 64×128×256 | 2×2 | (0,0) | 2 | --- |
| | Up_Conv_1_1 | 128×128×256 | 64×128×256 | 3×3 | (1,1) | 1 | ReLU |
| | Up_Conv_1_2 | 64×128×256 | 64×128×256 | 3×3 | (1,1) | 1 | ReLU |
| Out_ConvBlock | Out_Conv | 64×128×256 | 2×128×256 | 1×1 | (0,0) | 1 | --- |

*Abbreviations:* ConvGRU, convolutional gated recurrent unit; ConvLSTM, convolutional long short-term memory; SCNN, spatial convolutional neural network; ST-RNN, spatial-temporal recurrent neural network; ReLU, Rectified Linear Unit.

* Similar to the SegNet-based network architecture, two types of ST-RNN, i.e., ConvLSTM and ConvGRU, are tested;
** ST-RNN blocks are tested with one hidden layer or two hidden layers.